\documentclass{article}

\PassOptionsToPackage{numbers, compress}{natbib}

    \usepackage[preprint]{neurips_data_2024}

\usepackage[utf8]{inputenc} 
\usepackage[T1]{fontenc}    
\usepackage{hyperref}       
\usepackage{url}            
\usepackage{booktabs}       
\usepackage{amsfonts}       
\usepackage{nicefrac}       
\usepackage{microtype}      
\usepackage{xcolor}         

\usepackage{amsmath}
\usepackage{graphicx}
\usepackage{makecell}
\usepackage{enumitem}
\usepackage{color, colortbl}
\usepackage{tikz}
\usepackage{subcaption}
\usepackage{wrapfig}
\usepackage{multirow}
\usepackage{listings}
\usepackage{algorithm}
\usepackage{algorithmic}
\usepackage{stfloats}
\usepackage{xspace}
\usepackage{cleveref}

\crefname{section}{Sec.}{Secs.}
\Crefname{section}{Section}{Sections}
\crefname{table}{Tab.}{Tabs.}
\Crefname{table}{Table}{Tables}
\crefname{figure}{Fig.}{Figs.}
\Crefname{figure}{Figure}{Figures}
\crefname{equation}{Eq.}{Eqs.}
\Crefname{equation}{Equation}{Equations}

\definecolor{tabhighlight}{HTML}{e5e5e5}

\newcommand{\method}{{\color[RGB]{0,0,0}\text{StyleBooth}}\xspace}
\newcommand{\styletuner}{{\color[RGB]{0,0,0}\text{Style Tuner}}\xspace}
\newcommand{\styletuners}{{\color[RGB]{0,0,0}\text{Style Tuners}}\xspace}
\newcommand{\destyletuner}{{\color[RGB]{0,0,0}\text{De-style Tuner}}\xspace}
\newcommand{\destyletuners}{{\color[RGB]{0,0,0}\text{De-style Tuners}}\xspace}

\title{StyleBooth: Image Style Editing with \\Multimodal Instruction}

\author{Zhen Han, Chaojie Mao, Zeyinzi Jiang, Yulin Pan, Jingfeng Zhang \\
  Tongyi Lab, Alibaba Group. \\
  \tt \{hanzhen.hz,chaojie.mcj,zeyinzi.jzyz,yanwen.pyl, \\ 
  \tt zhangjingfeng.zjf\}@alibaba-inc.com \\
}
  
\begin{document}

\maketitle

\begin{abstract} \label{sec:abs}


Given an original image, image editing aims to generate an image that align with the provided instruction. 
The challenges are to accept multimodal inputs as instructions and a scarcity of high-quality training data, including crucial triplets of source/target image pairs and multimodal (text and image) instructions. 
In this paper, we focus on image style editing and present StyleBooth, a method that proposes a comprehensive framework for image editing and a feasible strategy for building a high-quality style editing dataset. 
We integrate encoded textual instruction and image exemplar as a unified condition for diffusion model, enabling the editing of original image following multimodal instructions. 
Furthermore, by iterative style-destyle tuning and editing and usability filtering, the StyleBooth dataset provides content-consistent stylized/plain image pairs in various categories of styles. 
To show the flexibility of StyleBooth, we conduct experiments on diverse tasks, such as text-based style editing, exemplar-based style editing and compositional style editing. 
The results demonstrate that the quality and variety of training data significantly enhance the ability to preserve content and improve the overall quality of generated images in editing tasks. 
Project page can be found at \url{https://ali-vilab.github.io/stylebooth-page/}.

\end{abstract}

\section{Introduction} \label{sec:intro}

\begin{figure}
  \centering
  \includegraphics[width=1.0\linewidth]{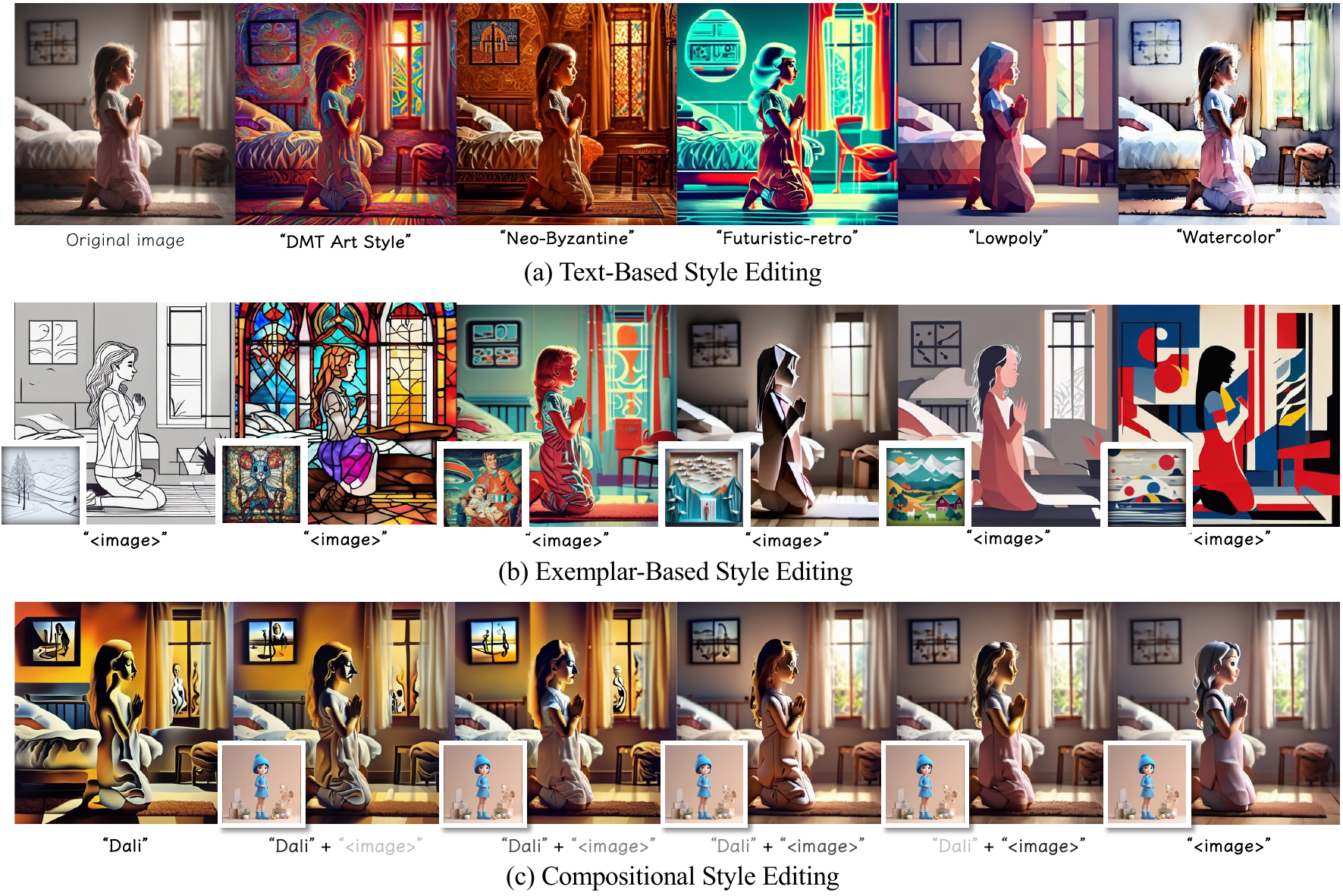}
  \caption{\textbf{Edited images by \method.} Based on multimodal instructions, \method supports 3 types of image style editing. Following the same instruction template: "Let this image be in the style of <style>/<image>", we conduct (a) text-based style editing and (b) exemplar-based style editing (c) compositional style editing. "<style>" or "<image>" is the identifier of textual style name and visual exemplar images. The style name is placed under the result image and exemplar is shown at the left-bottom corner. In (c), the style name and identifier "<image>" are marked in different color fading levels. The degree of fading represents the proportion of the corresponding style in the result images. Tuned by our elegantly designed style editing data, \method is capable of generating high-quality output images in diverse styles.}
  \label{fig:head}
  \vspace{-15pt}
\end{figure}



Diffusion models\cite{ddpm,ddim,ldm} have demonstrated impressive capabilities in Text-to-Image (T2I) generation. Image editing, with its extensive applications in everyday life, aims to modify original images to meet specific criteria given the reference image or prompt. Recently, diffusion-based image editing has emerged as a field with significant potential. Numerous prior studies have leveraged diffusion models to manipulate original images according to instructions. These methods include manipulating features of attention mechanism\cite{p2p,masa}, implementing guided diffusion during the denoising step\cite{selfguidance,dragondiff,uniguidance}, and tuning T2I models using image pairs for supervision\cite{zero123,phd}, and so on. 
However, these methods face the common challenges: \emph{(1)} only support instruction in a single modality (text or image), \emph{(2)} insufficient supervision in general image editing scenarios, particularly ignoring the effect of high-quality images and crucial pairs comprising target images and multimodal (text and image) instructions.



Image editing encompasses overall seven distinct tasks, as outlined in \cite{emuedit}, including style editing, object removal, object addition, background modification, color and texture alterations, local adjustments, and comprehensive global changes. It's a significant challenge to create an instruction-based image editing dataset that encompasses all these tasks, featuring high-quality images paired with comprehensive instructions. In this study, we focus on image style editing, which is a common and extensively employed aspect of image editing endeavors. Style transfer is a quintessential task in image-to-image (I2I) editing, with many efforts primarily directed at constructing style-specific models\cite{scenimefy} that support a limited range of styles, or at developing text-guided style editing models\cite{sdedit} that interpret textual descriptions of styles as inputs. To overcome the constraints imposed by textual style descriptions, exemplar-based style editing approaches\cite{agilegan, inst,vct} have been introduced, which accept any given style image as an additional input. However, these methods often sacrifice the capability to adhere to textual instructions.

To tackle these challenges, we introduce \method, a unified method for text and exemplar-based style editing that accommodates hybrid editing instructions combining written text and exemplar images. Concretely, we devise a unified conditioning schema for the diffusion model, using the tokens "<style>" and "<image>" as identifiers to indicate arbitrary style and exemplar style images. The text instruction, which substitutes the "<style>" token with specified style words, is processed through the text encoder. 
Additionally, the exemplar image is encoded by the image encoder, and the image feature is aligned with the encoded text feature in a unified hidden space, forming the multimodal input for the diffusion model. Merging the flexibility of natural language with the explicitness of exemplar images enables the exploration of more creative and personalized styles, such as style interpolation and style composition.
We present StyleBooth models learning from content-strong-aligned and high-resolution image pairs, which is capable of generating high content fidelity and quality edited images. Samples of edited results are shown in ~\cref{fig:head} and we also present a high-quality style editing dataset, constructed through an effective pipeline combined iterative edit tuning and usability filtering. This strategy is easy to generalize to any other image editing tasks. 

In summary, this work introduces a comprehensive framework for image editing that supports multimodal instructions, along with a meticulously crafted data generation pipeline that assembles a high-quality style editing dataset. We carry out experiments across a range of editing tasks, including text-based style editing, exemplar-based style editing, and compositional style editing. The results reveal that the quality and stylistic diversity of training data markedly improve content preservation and elevate the overall quality of the images produced during the editing process.

\section{Related work} \label{sec:related}

\textbf{Image style transfer.}
Image style transfer has been widely studied by researchers. It involves the process of applying stylistic representations to content images in order to generate images with specific styles.
Style transfer based on deep neural networks ~\cite{styletrans,controlstyletrans} typically employs separating and recombining the content and style of an image to achieve a simple transfer of style, but there is a problem with the precision of the style representation. 
There are also works based on GANs ~\cite{pix2pix,cyclegan,stargan,stylegan} that, through training generators and discriminators, apply the representations of style images while preserving the features of content images as much as possible. However, these methods lack sufficient adaptability and the trained models need to correspond to the style, which limits the possibility of more extensive transformation. 
Recently, diffusion-based image generation has developed rapidly and has also been applied to style transfer tasks. These methods~\cite{inst,vct,styledrop,instantid} often require users to provide only a few or even just one style image to achieve a relatively universal style transfer.
In this process, generating high-quality transferred images while maintaining the semantic information of the content image remains a challenge.
Our \method builds upon this foundation and can continuously generate high-quality transferred images while preserving the semantic information of the content image.

\textbf{Text-based image editing.}
Image editing is a common application in practical scenarios, where it can achieve customized image generation through external guidance.
Many pre-trained generative models based on diffusion models~\cite{ddpm,ddim,sde,glide,ldm} can generate corresponding images according to text, but it is difficult to finely edit images according to specific needs while keeping the main content unchanged.
Recent works edit global semantic information of images based on text guidance~\cite{text2live,p2p,nullinv}. However, they still require detailed descriptions or corrections of the target images.
Text-based image editing offers great convenience, as it only requires describing the specific changes to be made to the image, such as \textit{"Let it be in the style of Picasso"}.
InstructPix2Pix~\cite{ip2p} is trained with instructional data generated by GPT-3~\cite{gpt3} and Prompt-to-Prompt~\cite{p2p}. 
MagicBrush~\cite{magicbrush} builds an image editing dataset with manually annotated instructions using online image editing tools.
Emu Edit~\cite{emuedit} proposes a multi-task image editing model and trains it on image editing and recognition data.
Although these works have achieved improvements, high-quality data and precise execution of editing instructions remain significant challenges.

\textbf{Exemplar-based style editing.}
Exemplar-based image editing typically uses one or several image exemplars as references for generations.
InST~\cite{inst} introduces an attention-based textual inversion method that inverts a painting into corresponding text embeddings to guide the model in creating images with a specific artistic appearance.
StyleDrop~\cite{styledrop} employs an adapter fine-tuning approach to efficiently train models corresponding to the reference images.
RAVIL~\cite{rival} modifies the denoising inference chain of diffusion models to align with the real image inversion chain, generating diversified and high-quality variants of real-world images.
The aforementioned methods only support editing with exemplar images, but for more detailed guidance, our \method can generate highly creative images based on any image exemplar and text instruction while maintaining the consistency of the identity of the images to be edited.

\begin{figure}[t]
  \centering
  \includegraphics[width=1.0\linewidth]{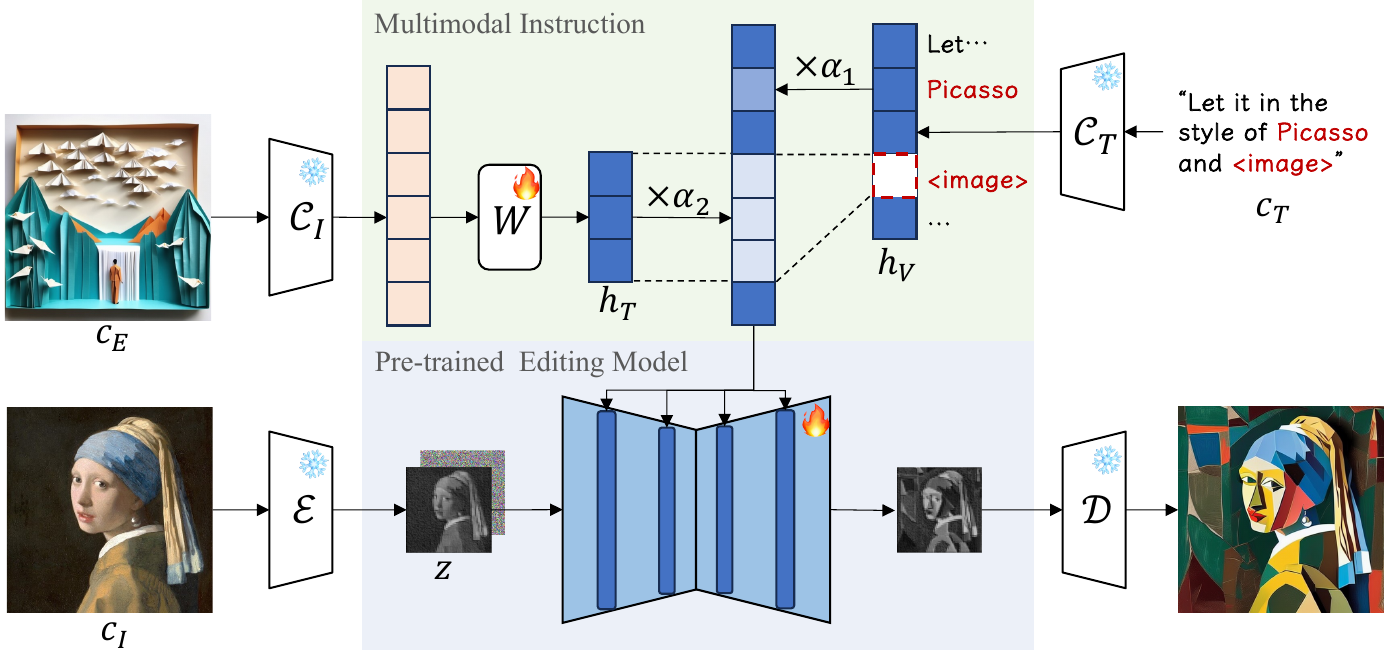}
  \caption{\textbf{Overview of \method method.} We propose Multimodal Instruction, mapping the text input and exemplar image input into a same hidden space through a trainable matrix $W$, which unifies vision and text instructions. 
  The textual instruction templates are carefully designed, introducing undetermined identifiers like "<style>" and "<image>" to support multimodal inputs. 
  To balance every style for compositional style editing, we conduct Scale Weights Mechanism $\alpha_i$ on the hidden space embeddings.
  Editing is conditioned by multimodal features following the composed instructions from different modalities at the same time.
  }
  \label{fig:method2}
  \vspace{-10pt}
\end{figure}

\section{StyleBooth} 
\label{sec:method}


In this section, we present a unified text and exemplar-based style editing method namely \method.
For styles easy to be explained in text, the text-based method is capable of following text instructions. 
As for those hard to write accurately, one may use an exemplar-based method providing the target style by uploading a reference image. 
\method has both advantages. 
\method unifies text instruction and exemplar image into a multimodal instruction.
Our model is a latent diffusion model conditioned by this multimodal instruction together with an image to be edited as illustrated in~\cref{fig:method2}.
High-quality data is also essential to style editing tasks. 
We introduce a paired image constructing strategy based on iterative style-destyle editing and usability control.
Following this methodology, we present a high-quality dataset supporting text and exemplar-based style editing.

\subsection{Multimodal Instruction Editing} 

\textbf{Preliminaries.}
Diffusion models\cite{ddpm} are trained to sample image $x$ from Gaussian noise through a reversed diffusion process, which is a sequence of denoising operations of predicting the added noise $\epsilon\sim\mathcal{N}(0,1)$ at a specific timestamp $t$.
To improve the efficiency, latent diffusion\cite{ldm} utilizes a pre-trained variational autoencoder\cite{vae} $\mathcal{E}$ to encode $x$ into latent space $z=\mathcal{E}(x)$ and operates denoising in the compressed latent space.
InstructPix2Pix\cite{ip2p}, an image editing model, tuned from a pre-trained T2I latent diffusion model $\epsilon_\theta$ conditioned by instruction text $c_T$ and original image $c_I$. 
$c_T$ is encoded by pre-trained CLIP\cite{clip} text encoder into text features  $h_T=\mathcal{C}_T(c_T)$, while $c_I$ is encoded by a pre-trained variational autoencoder into latent.
The training objective is formulated as follows:
\begin{align}
  L = \mathbb{E}_{\mathcal{E}(x),\mathcal{E}(c_I),h_T,\epsilon,t}\Vert\epsilon_\theta(z_t, t,\mathcal{E}(c_I),h_T)-\epsilon\Vert_{2}^{2}.
\end{align}

\textbf{Multimodal instruction.} \label{sec:method_ins}
To support text-based and exemplar-based style editing at the same time, our model supports text-only, image-only, and text-image instructions, as illustrated in~\cref{fig:method2}. 
Input images are treated as an undetermined identifier "<image>" in the text encoder.
We encode the style exemplar image $c_E$ using CLIP image encoder $\mathcal{C}_I$. The patch features after the last Transformer layer are considered as exemplar image features.
We introduce a trainable convolutional alignment layer $W$ to map image features into text feature space:
\begin{align}
  h_V = W\mathcal{C}_I(c_E).
\end{align}
Finally, replacing the feature corresponding to the identifier "<image>", aligned $h_V$ are inserted into $h_T$ as visual tokens. The final multimodal instruction is formulated as:
\begin{align}
  h = \mathcal{F}_{insert}(h_T, h_V).
\end{align}

\textbf{Textual instruction templates for multimodal inputs.}
For diverse, comprehensible, and grammar-correct style editing instructions, we leverage the Turbo 128k version of GPT4\cite{gpt4} to generate editing instructions. 
Instead of complete sentences, we ask LLM for literal templates composed of normal text and identifier tokens. These identifiers represent the undetermined target styles. 
We write examples to make sure LLM to precisely interpret our intent.
For text-based style editing, we give an instruction example "Let this image be in the style of <style>".
For exemplar-based style editing, we use "Let this image be in the style of <image>" as an instruction example.
Notice that we use "<style>" and "<image>" as the identifiers of arbitrary styles in types of text and exemplar.
Adequate templates are generated for each task for training.
During inference, input instructions can be varied and unnecessarily limited to the training instructions.

\textbf{Text-based and exemplar-based style editing.} \label{sec:method_edit}
Text-based methods perform editing following text-only instructions.
During training, instruction templates are randomly selected and the identifier "<style>" is replaced by the actual style name. 
We follow the methodology of InstructPix2Pix and fine-tune from their pre-trained model. Supervised by the paired images, the model learns to add the specific style following text-only instructions.

Style of exemplar image and that of target image must be well-aligned during training phase. 
We search for the most appropriate exemplars from other target images in the same style subset.
Within a style subset, we first calculate CLIP score for every target image with other in-class stylized images. Images with the highest score are the most similar ones and they are the most appropriate candidates as exemplar-style images. During training, the final exemplar will be randomly selected from the top 10 most similar images.
An instruction template is randomly selected and encoded into text features by $\mathcal{C_T}$, where the identifier "<image>" is treated as an unknown identifier at the beginning. Composed of exemplar image features, the original textual instruction is transformed into a multimodal instruction $h$.
The training objective is changed into:
\begin{align}
  L = \mathbb{E}_{\mathcal{E}(x),h,\mathcal{E}(c_I),\epsilon,t}\Vert\epsilon_\theta(z_t, t, h,\mathcal{E}(c_I))-\epsilon\Vert_{2}^{2}.
\end{align}

\textbf{Scale weighting mechanism for compositional style editing.} \label{sec:method_scale}
As shown in~\cref{fig:method2}, in order to balance the presence of visual elements, we introduce a Scale Weighting Mechanism.
For $n$ textual or visual style representations, we assign a series of multiplicative scale factors $\alpha_1,\alpha_2,\ldots,\alpha_n$. We find the Scale Weighting Mechanism effective in controlling the style presence ratio in the final results.

\begin{figure}[t]
  \centering
  \includegraphics[width=1.0\linewidth]{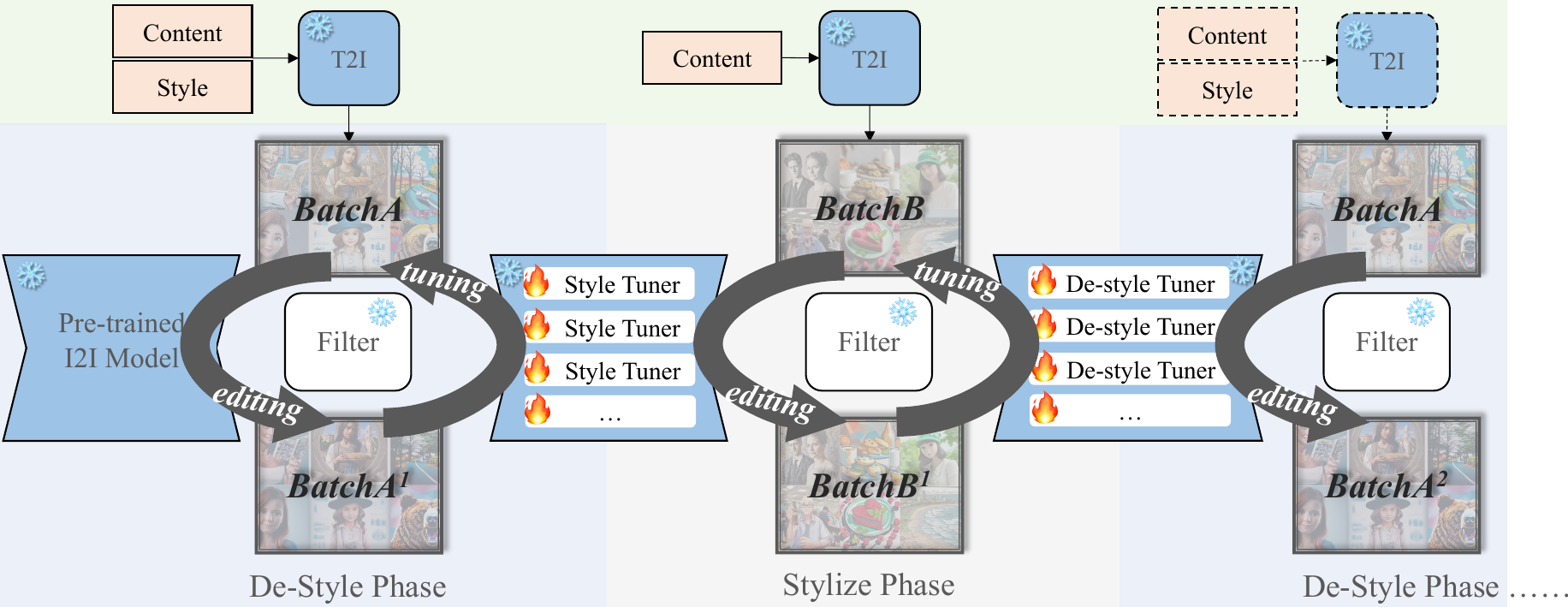}
  \caption{\textbf{Iterative Style-Destyle Tuning and Editing pipeline.} Following a de-style editing, filtering, style tuning, stylize editing, filtering and de-style tuning steps, Iterative Style-Destyle Tuning and Editing leverages the image quality and usability. 
  }
  \label{fig:method}
  \vspace{-10pt}
\end{figure}

\begin{figure}[!ht]
  \centering
  \includegraphics[width=0.95\linewidth]{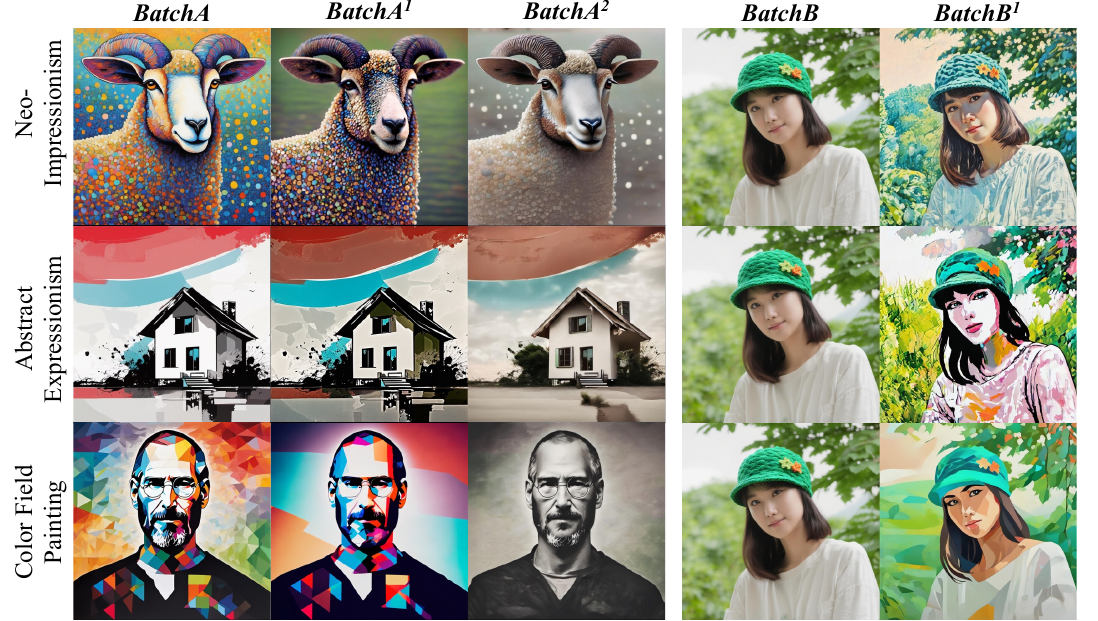}
  \caption{\textbf{Generated samples of the intermediate and final image pairs during Iterative Style-Destyle Tuning and Editing.} During iterations, image quality gets higher while key style features are gradually wiped off in the de-styled images. We show the style images and de-style results generated in 1st and 2nd de-styled phase and a plain image and results generated in 1st stylize phase.
  }
  \label{fig:dataset}
  \vspace{-10pt}
\end{figure}

\subsection{High-Quality Style Editing Dataset} \label{sec:method_data}
We aim to obtain a tuning dataset for style editing \emph{(1)} covering a broad category of styles, \emph{(2)} consisting of stylized/plain image pairs with identical content and \emph{(3)} made up of entirely high-quality images. 
The key idea we constructing such a dataset is starting from high-quality stylized and plain unpaired images, then conducting a basic I2I style adding and removal transformation to generate image pairs as intermediate training data.
We improve the transformation quality by a filter mechanism during multi-round Iterative Style-Destyle Tuning and Editing. 

We take advantage of the T2I ability of an advanced T2I diffusion model and the ability of an text-based image editing model pre-trained from the InstructPix2Pix\cite{ip2p} dataset. 
Instead of collecting from real images, we use T2I-synthesized images for convenience and avoiding possible moral violations.
To leverage the quality of generated image pairs, translation between style image and plain image is done iteratively and usability filter is conducted after each iteration. 
The pipeline of data generation is illustrated in~\cref{fig:method}.

\textbf{Generating style and plain images.} 
Intuitively, images in specific styles can be generated by any T2I model as long as the style-related keywords to the text prompt. 
We follow the same route but design text prompts carefully through prompt expansion provided by Fooocus\cite{fooocus} to avoid the style of generated image misaligned with our prompts. 
For each of the styles, prompts are modified by the style format before being sent to T2I model. 
We accept 67 prompt formats and 217 diverse content prompts, ending up with 67 different styles and 217 images per style. This batch of stylized images is marked as $BatchA$.
Next we generate $BatchB$ consisting of plain images.
For diversity and quality of generated images, we randomly select image captions from the LAION Art\cite{laion5b} dataset as prompts. Unlike style image generation, no specific style prompt expansion is used.

\textbf{Vanilla de-style and usability filtering.}
We train a vanilla version of image editing model following the methodology of Instruct-Pix2Pix\cite{ip2p}. 
This model has a basic I2I editing ability but the generated data quality falls far short of our needs. 
Starting from $BatchA$, we use the image editing model to de-style them into plain photographic images generating the corresponding $BatchA^1$. 
As shown in the second column of~\cref{fig:dataset}, after the first de-style operation, the content structure is preserved yet most of these de-styled images contain features of the original style. We then employ a CLIP-based\cite{clip} metric to filter out image pairs that are too similar, indicating no significant style change, and that are overly different, which could imply potential content distortion. By establishing upper and lower thresholds for CLIP scores, this filtering process effectively eliminates the majority of evident failures produced by the image editing model.

\textbf{Iterative style-destyle tuning and editing.}
To further strengthen the pre-trained I2I model, we take filtered image pairs as training data and fine-tune it using \styletuner to learn style-adding ability. 
Now, the $BatchA^1$ are used as inputs while the style images in $BatchA$ as targets.
Efficient tuning methods\cite{lora, restuning, controllora, scedit} show strong few-shot learning capability.
Here, we add efficient tuners called \styletuners to the basic I2I model and tune different styles separately which means each tuner is only responsible for one specific style.
After tuning, the base model with tuners will edit $BatchB$ images to add the styles they have learned. Then we get $BatchB^1$.
Some of the results are shown in the last column of~\cref{fig:dataset}. The style-adding results is frustrating and styles in $BatchB^1$ is still not that accurate. 
Again, $BatchB$ and $BatchB^1$ images are measured and filtered by CLIP-based metric to be the next round training data.
Note that we always use the high-quality images in $BatchA$ or $BatchB$ as the training targets.
In this round, we tune I2I model to be a style removal model using \destyletuner, which means images in $BatchB^1$ are used as inputs and image in $BatchB$ as targets. 
Still, one tuner is trained to remove one type of style. 
The tuned I2I model edits the stylized image in $BatchA$ and geneates $BatchA^2$ which are shown in the third row of~\cref{fig:dataset}. The image quality in $BatchA^2$ is improved dramatically.
We also use CLIP scores to filter out unqualified image pairs.
After $n$ iterations, $BatchA^n$ are generated and image pairs of $BatchA$ and $BatchA^n$ are filtered. 
The style-diverse, high-quality and content-preserved stylized/plain image pairs are finally observed. 
We compare the number of images those passing the CLIP filter.
Comparing to the results of vanilla de-style, the average usability rate is increased dramatically from 38.11\% to 79.91\%
by 41.80\%, which shows that our Iterative Style-Destyle Editing is effective to improve paired images' quality.

\section{Experiments} \label{sec:exp}

\subsection{Experimental setup}
We conduct our data generation and experiments on an open-source project SCEPTER\cite{scepter}, integrating many styles from Fooocus\cite{fooocus} and providing an easy implementation of efficient tuning methods.
We use SDXL\cite{sdxl} to generate $BatchA$ and $BatchB$. LoRA\cite{lora} with rank 256 are accepted as \styletuners and \destyletuners. The editing-tuning iterations stop at the end of the 2nd de-style process. So the final training pairs are $BatchA^2$ and $BatchA$.
The trainable $W$ is configured as a $6\times6$ convolutional layer with stride 4, mapping a $14\times14$ visual features into 9.

\begin{table}[tb]
  \caption{CLIP-based metrics and user study on Emu Edit benchmark. For each baseline, we report CLIP similarities and success rate of instruction-following, content-preserving. For preference rate, we show the percentage of samples that raters prefer the results of \method.
  }
  \label{tab:metrics}
  \centering
  \setlength{\tabcolsep}{8pt}
  \begin{tabular}{@{}lcccccc@{}}
    \toprule
    Methods & CLIP$_{dir}$ & CLIP$_{out}$ & CLIP$_{img}$ & SR$_{ins}$ (\%)& SR$_{cont}$ (\%)& UPR (\%) \\
    \midrule
    InstructPix2Pix & \textbf{0.1082} & 0.2271& 0.7251  & 76.55 & 90.04 & 50.44 \\
    MagicBrush      & 0.1060 & 0.2202& 0.7174  & 61.95 & 82.30 & 63.27  \\
    Emu Edit        & 0.1006 & \textbf{0.2378}& \textbf{0.7552}  & 76.55 & \textbf{97.12} & 58.85 \\
    Ours            & 0.1062 & 0.2293& 0.7030  & \textbf{77.65} & 94.47& -     \\
  \bottomrule
  \end{tabular}
  \vspace{-10pt}
\end{table}

\subsection{Text-Based Style Editing}

\textbf{Baselines and benchmarks.}
We compare our model with 3 state-of-the-art text-based image editing methods.
Both InstructPix2Pix\cite{ip2p}, Magic Brush\cite{magicbrush} and The Emu Edit\cite{emuedit} present machine-generated image editing instruction-tuning datasets involving triplets of the original image, edited image, and text instruction.
Magic Brush mainly focuses on local editing tasks while InstructPix2Pix and
Emu Edit are more generalized. Additionally, Emu Edit introduces an instruction-based image-editing benchmark. Each entry in this benchmark includes an input image, a set of instructions, captions for both the input and output images, and an edited result generated by the Emu Edit model. This benchmark encompasses seven distinct tasks: style editing, object removal, object addition, background alteration, color/texture modification, local adjustments, and comprehensive transformations. In our study, we concentrate exclusively on the style editing task, which comprises a total of 226 samples after excluding one sample with blank input and output images.

\textbf{Evaluation metrics.}
We use the following \textbf{CLIP-based metrics}: 
\emph{(1)} Directional score(CLIP$_{dir}$)\cite{stylegannada}  measures the alignment of 2 directions, input-output caption change direction and input-output image change direction in CLIP space. The input-output captions are provided in Emu-Edit Benchmark.
\emph{(2)} Image similarity(CLIP$_{img}$) measures the similarity between the input and edited images indicating the number of changes.
\emph{(3)} Output similarity(CLIP$_{out}$) measures the similarity between the edited image and the output caption. 
To assess editing outputs under human standards, three \textbf{human rating scores} are presented: 
\emph{(1)} Instruction success rates(SR$_{ins}$) estimates whether output styles are correct and match instructions.
\emph{(2)} Content success rate(SR$_{cont}$) estimates the content consistency.
\emph{(3)} User preference rate(UPR) compares our method against other baseline approaches.

\textbf{Quantitative comparisons.}
As displayed in~\cref{tab:metrics}, our method achieves competitive CLIP$_{dir}$ and CLIP$_{out}$ scores in comparison to the previously mentioned baselines. Emu-Edit achieves significantly higher CLIP$_{img}$ and SR$_{cont}$ scores than others, which means the outputs and the original images are more similar in both CLIP space and human views. It's essential to measure content-preservation, but an extremely high score might come from insufficient conveying of the style features. We will explain more in qualitative comparison.
Overall, our approach attains the highest instruction-following rate SR$_{ins}$ at 55.44\% and an elevated user preference rate UPR of 58.85\%, outperforming the baseline methods. These figures indicate that our model not only has the highest rate of accurate edits but is also most favored by users.

\begin{figure}[t]
  \centering
  \includegraphics[width=1.0\linewidth]{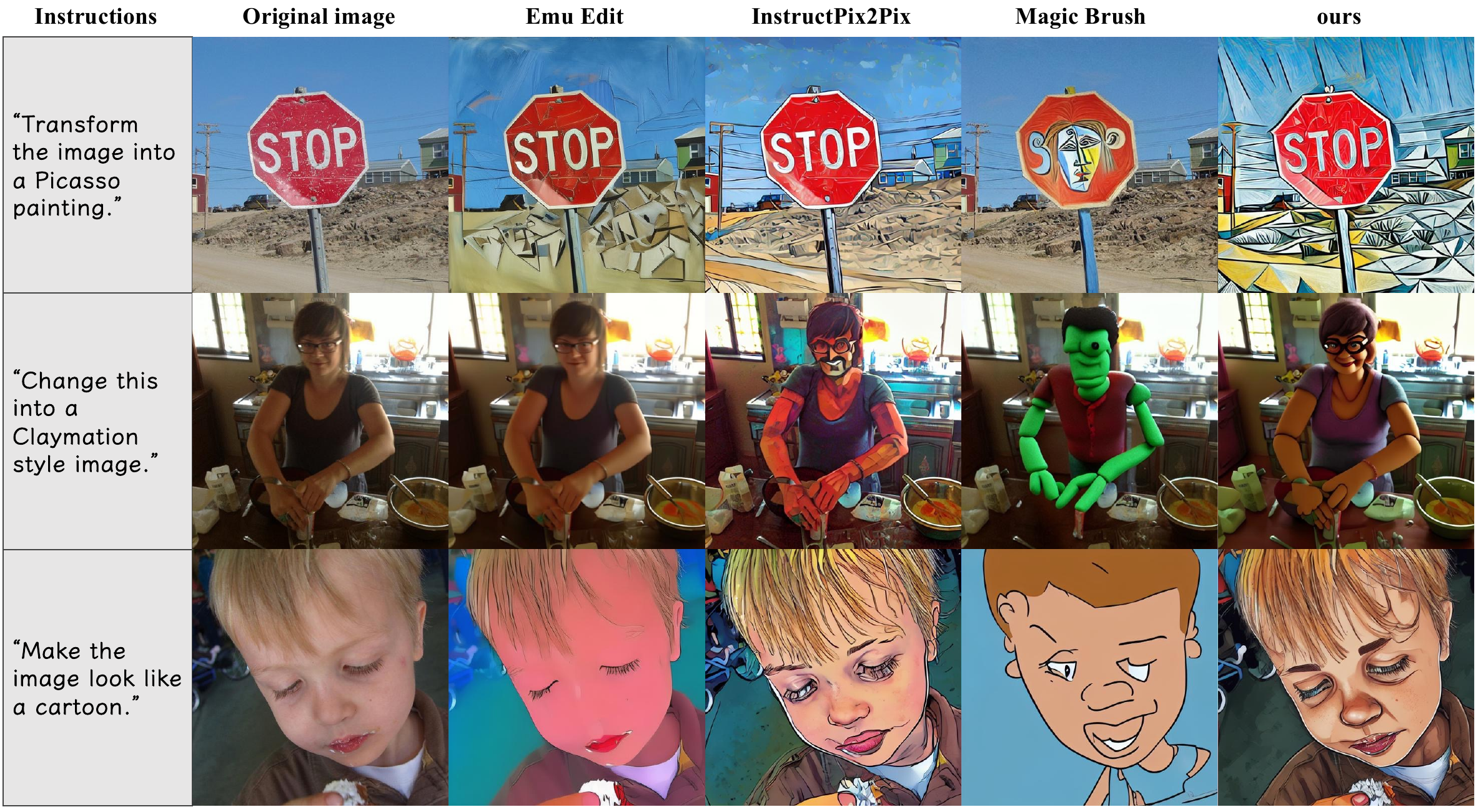}
  \caption{\textbf{Comparisons with instruction-based style editing baselines in Emu Edit benchmark.} We show editing results of \method and 3 baselines. The results of \method are the most accurate in both style conveying and content preservation comparing to others, though some of the styles and instruction syntax are not contained in our tuning dataset.
  }
  \label{fig:bench}
  \vspace{-10pt}
\end{figure}

\textbf{Qualitative comparisons.}
 The editing results of \method alongside those of the baselines are shown in ~\cref{fig:bench}. Our approach exhibits commendable performance with novel styles, such as "Picasso" and "Claymation", as well as with instructions framed in previously unseen syntax, demonstrating robust zero-shot learning abilities in text-based tasks. As previously discussed, Emu Edit falls short in adequately applying the requisite style features to the edited images, leading to a high resemblance between the original and the edited images.
Additionally, Magic Brush has a propensity to concentrate on local areas after fine-tuning on datasets specific to local editing, even when engaged in style editing tasks. In comparison to InstructPix2Pix, our method displays superior precision in style editing and mitigating artifacts. Above all, our edited images stand out as the most accurate and visually appealing among those produced by competing methods.

\subsection{Exemplar-Based Style Editing}
\textbf{Baselines.}
We conduct comparisons with three state-of-the-art methods capable of extracting styles from image exemplars. Both RIVAL\cite{rival} and StyleAligned\cite{stylealigned} are inversion-based methods inverting exemplar images into an initial noisy latent space then infusing features via attention mechanisms. In the VCT\cite{vct} approach, both the exemplar and source images are optimized into a latent space, where style and content are fused via classifier-free guidance\cite{cfg} and attention. It should be noted that RIVAL and StyleAligned are originally proposed for exemplar-based text-to-image generation task. Therefore, we give a textual prompt derived from the original image to provide the necessary content information. To further enhance their content preservation, we also include results with Canny-based\cite{canny} and depth-based\cite{depth} ControlNet\cite{controlnet}, as utilized in their official implementations.

\begin{figure}[t]
  \centering
  \includegraphics[width=1.0\linewidth]{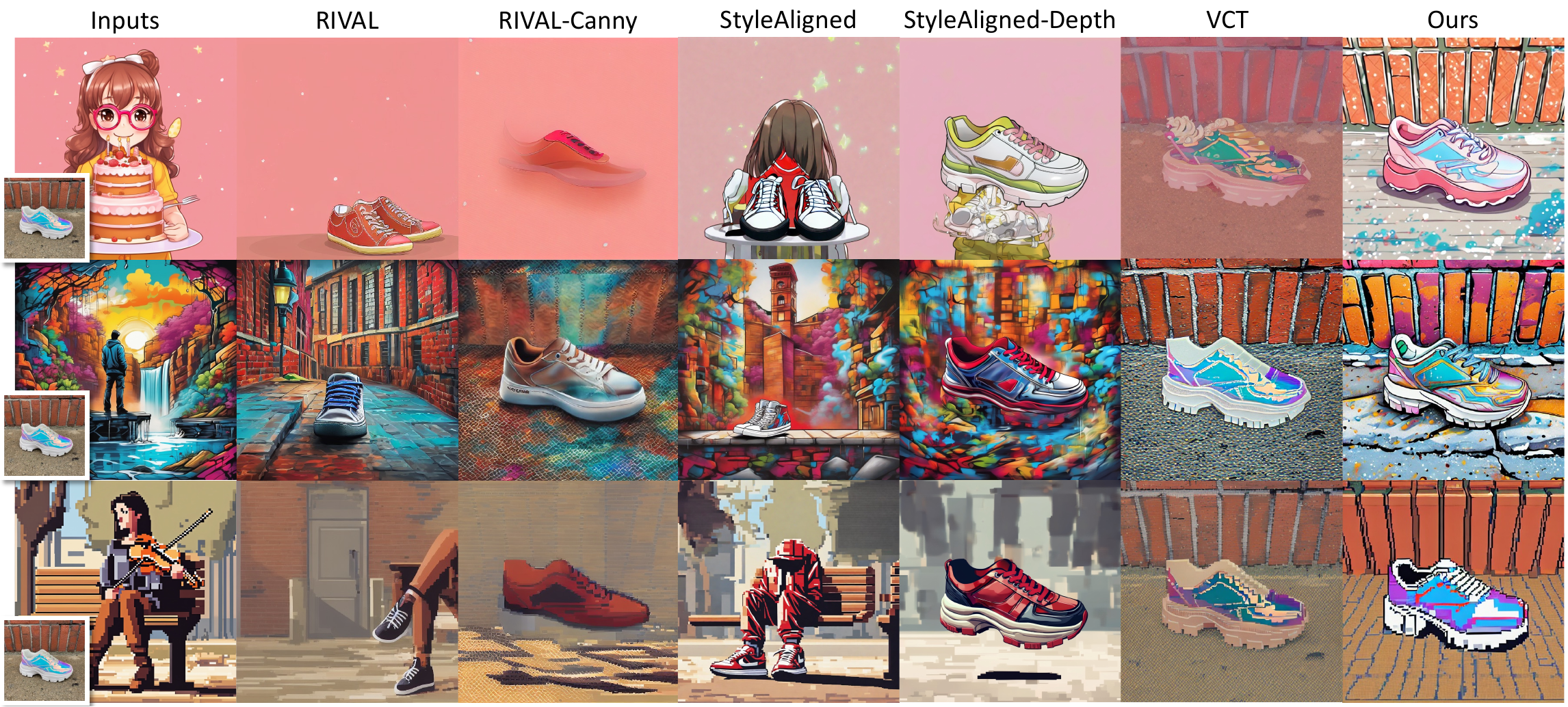}
  \caption{\textbf{Exemplar-based style editing comparisons} 
  The inputs include a style exemplar and an original image displayed on the left-bottom corner of exemplar images. \method achieves both accurate style extraction and transformation and identical content preservation.
  }
  \label{fig:exemplar}
  \vspace{-10pt}
\end{figure}

\textbf{Comparisons.}
Exemplar-based style editing comparisons are displayed in ~\cref{fig:exemplar}. Both RIVAL and StyleAligned produce images that effectively capture the style features of the exemplars. However, they struggle to keep content-consistent to original image, even when combined with the Canny-based or depth-based ControlNet. Optimizing a content embedding for each original image, VCT preserves a majority of the content information. Nevertheless, VCT falls short in terms of both style transfer precision and output image quality. In summary, our method not only supports exemplar-based image style editing but also produces results that are both accurate and of a high aesthetic quality. Notably, our technique precisely extracts style features from the exemplars while faithfully maintaining the original image content for the style editing process.

\section{Style Composition and Interpolation.} \label{sec:comp_inter}
\begin{figure}[t]
  \centering
  \includegraphics[width=1.0\linewidth]{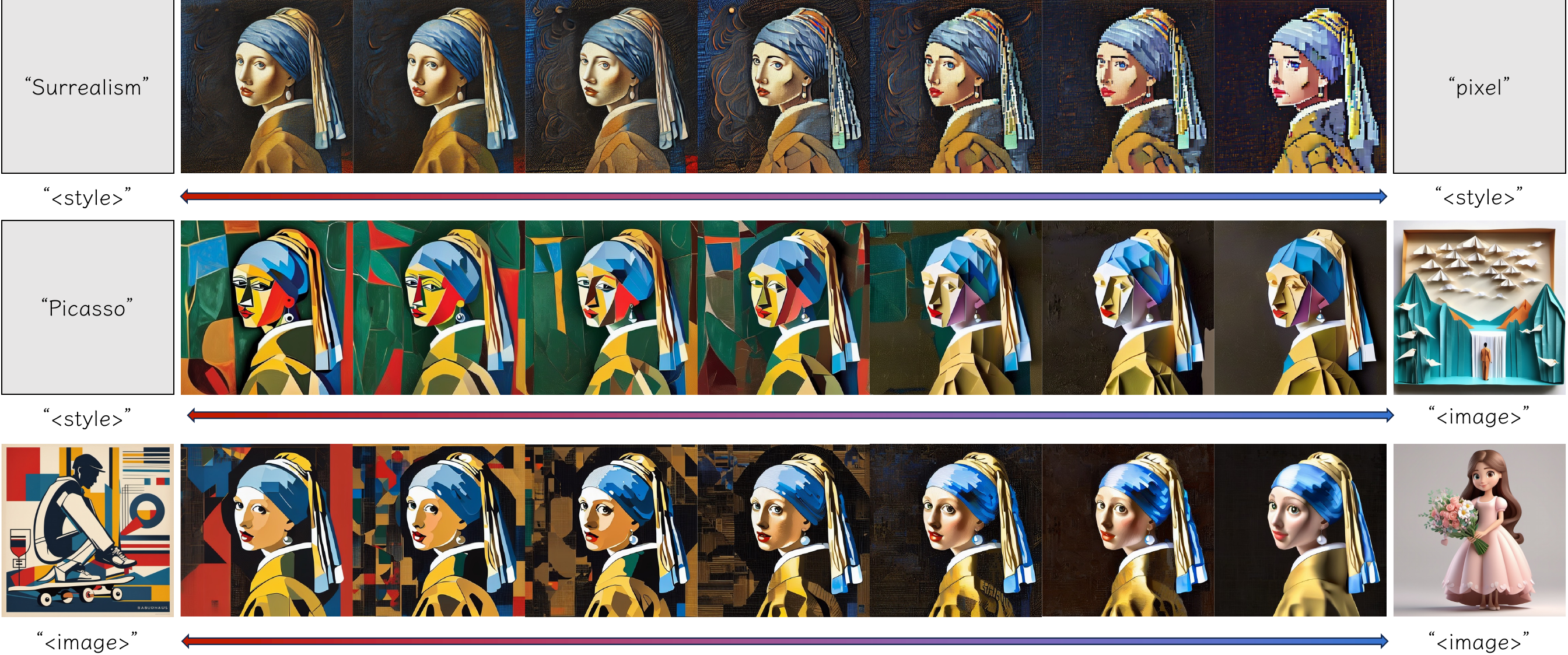}
  \caption{\textbf{\method compositional style editing and style interpolation.} \method unifies textual and visual exemplar by mapping them into a same hidden space making it possible to adjust the proportion of different styles in different modalities.}
  \label{fig:inter}
  \vspace{-10pt}
\end{figure}

We carry out style interpolation by manipulating the style scale weights $\alpha_i$. In the case of text-to-text and text-to-exemplar interpolation, we progressively reduce $\alpha_0$ from 1.5 to 0.5 while simultaneously increasing $\alpha_1$ from 0.5 to 1.5. For image-to-image interpolation, we blend the features of the two images into a single representation by adjusting a pair of weights, ensuring their sum remains normalized to one. As illustrated in~\cref{fig:inter}, by mapping both the text and the exemplar image into a unified hidden space, our multimodal instruction successfully narrows the gap between different modalities, thereby enabling the composition and blending of styles across various modalities.

\section{Conclusion} \label{sec:con}

We present \method as a multimodal instructional image style editing method. 
It independently encodes the reference image and text, subsequently transforming and aligning them within a latent space, followed by injection into the backbone network for generative guidance to achieve text- and exemplar-based instructional editing. 
Meanwhile, \method can also fuse multimodal information for a compositional creative generation. 
Furthermore, we construct a high-quality dataset for style editing, composed of a wide variety of content-consistent stylized and plain image pairs, which assists us in building better editing models.

\noindent \textbf{Limitations.} In this work, we construct a rich dataset for style editing. However, the data construction is based on the textual descriptions of specific styles, such as watercolor, which significantly limits the number of styles. Collecting more extensive editing datasets will be our future work.

{\small
\bibliographystyle{abbrvnat}
\bibliography{main}
}

\newpage

\appendix


\section{Overview}
In the appendix, we present more implementation details of StyleBooth Dataset in~\cref{sec:supp_data} including image pairs and textual instruction templates. We also provide the evaluation of our training data in~\cref{sec:supp_eval}. Secondly, we explain the experiment implementation details in~\cref{sec:exp_detail}. We show our model's style composition and interpolation ability and more results in~\cref{sec:comp_inter} and ~\cref{sec:supp_exp}. Furthermore, we discuss social impacts and licence of assets in~\cref{sec:discus}.
Finally, we explain the details of human evaluation in~\cref{sec:human_eval}.

We public the following materials:
\begin{table}[!h]
  \caption{Shared materials and licenses in this work.}
  \label{tab:material}
  \centering
  \begin{tabular}{@{}lp{10cm}l@{}}
    \toprule
    Material & Link & License \\
    \midrule
    Homepage        & \url{https://ali-vilab.github.io/stylebooth-page/} & - \\
    Code & \url{https://github.com/modelscope/scepter/tree/main/docs/en/tasks/stylebooth.md} &  Apache-2.0  \\
    Dataset         & \url{https://modelscope.cn/models/iic/stylebooth/file/view/master?fileName=datasets%252Fstylebooth_dataset.zip} & Apache-2.0  \\
    Model          & \url{https://modelscope.cn/models/iic/stylebooth/file/view/master?fileName=models%252Fstylebooth-tb-5000-0.bin} &  Apache-2.0  \\
  \bottomrule
  \end{tabular}
  \vspace{-15pt}
\end{table}

\section{Dataset Details.} \label{sec:supp_data}
\subsection{Style Categories.}

\begin{figure}[h]
  \centering
  \includegraphics[width=1.0\linewidth]{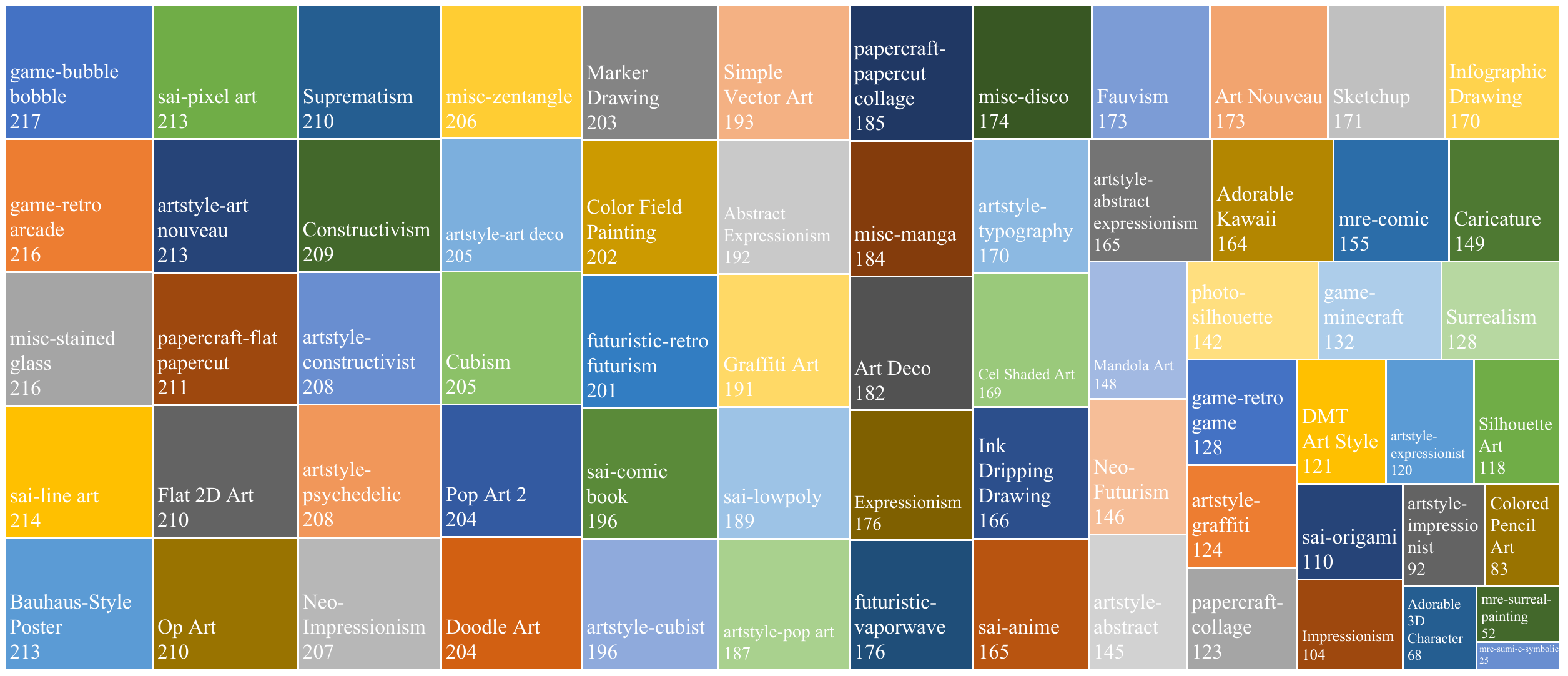}
  \caption{\textbf{Style distributions of StyleBooth dataset.} We present the name and image pair numbers for each style. Best viewed when zoomed in.}
  \label{fig:dist}
\end{figure}

The StyleBooth dataset consists of 63 style categories and the image pair numbers of each style are shown in ~\cref{fig:dist}. While some styles have fewer image pairs, the majority is relatively uniformly distributed with the number of image pairs being above 100.

\subsection{Image Pair Construction.}
\begin{figure}[tb]
  \centering
  \includegraphics[width=1.0\linewidth]{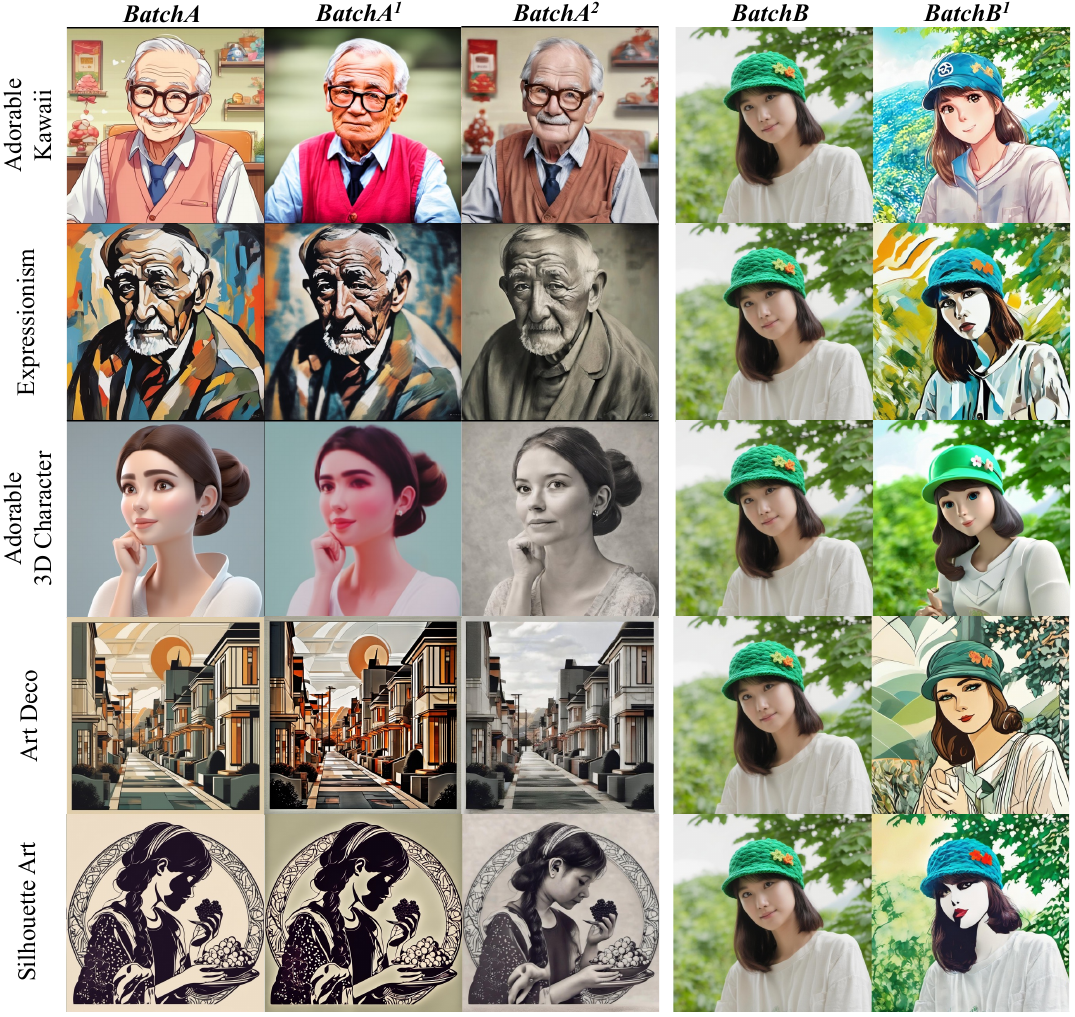}
  \caption{\textbf{Additional dataset samples generated in iterative style-destyle editing.} Pair $BatchA-BatchA^1$ and $BatchB-BatchB^1$ are intermediate image pairs, while final image pairs are $BatchA-BatchA^2$.
  }
  \label{fig:dataset2supp}
  \vspace{-10pt}
\end{figure}

We use stylize prompt expansion formats provided by Fooocus\cite{fooocus}, see the second column of~\cref{tab:styles} for examples. A placeholder of "\{prompt\}" is set for the original prompt.
To generate $BatchA$, we use 217 various prompts to generate diverse style images. 
Similarly, we select 200 image captions from LAION Art\cite{laion5b} dataset as prompts to generate $BatchB$.
5 samples of prompts used for $BatchA$ and $BatchB$ generation are shown in~\cref{tab:prompts}.
See the first column of~\cref{fig:dataset2supp} for $BatchA$ samples in different styles and the fourth column for $BatchB$.
To produce the derived images $BatchA^n$ and $BatchB^n$, we first train a vanilla version of image editing model using the Instruct-Pix2Pix\cite{ip2p} training data based on a pre-trained T2I diffusion model at 512$\times$512 resolution.
Each \styletuner and \destyletuner is tuned on 1 NVIDIA A800-SXM4-80GB GPU for 10000 steps. 
We use a learning rate of 0.0001 and a small batch size of 4 and the training resolution is set to 1024$\times$1024. Image pairs are randomly resized to $1\times$\ --\ $1.125\times$ training resolution and then center cropped into $1\times$. 
For usability filtering, we employ a CLIP-based metric and establish upper and lower thresholds of 0.84 and 0.2. From~\cref{fig:dataset2supp}, you can see the evolution of image quality comparing $BatchA^2$ to $BatchA^1$. The final data is the image pairs of $BatchA^2$ and $BatchA$ which is excellent as a style editing data.

\renewcommand\theadalign{cl}

\begin{table}[tb]
  \caption{Top 5 styles where the number of usable image pairs increase the most after iterative style-destyle editing. Numbers are reported in percentage(\%).
  }
  \label{tab:styles}
  \centering
  \setlength{\tabcolsep}{7pt}
  \begin{tabular}{@{}p{2.2cm}p{6cm}ccc@{}}
    \toprule
    Style Name & \multicolumn{1}{c}{Prompt Expansion Format} 
    & \multicolumn{1}{c}{$BatchA^1$} & \multicolumn{1}{c}{$BatchA^2$} 
    & \multicolumn{1}{c}{$\Delta$}\\
    
    \midrule
    artstyle-psychedelic & 
    "psychedelic style \{prompt\} . vibrant colors, swirling patterns, abstract forms, surreal, trippy"
    & 8.76 & 95.85 & 87.10\\
    Suprematism & 
    "Suprematism, \{prompt\}, abstract, limited color palette, geometric forms, Suprematism" 
    & 14.75 & 96.77 & 82.03\\
    misc-disco & 
    "disco-themed \{prompt\} . vibrant, groovy, retro 70s style, shiny disco balls, neon lights, dance floor, highly detailed"
    &  5.53 & 80.18 & 74.65\\
    Cubism & 
    "Cubism Art, \{prompt\}, flat geometric forms, cubism art"
    &  20.74 & 94.47 & 73.73\\
    Constructivism & 
    "Constructivism Art, \{prompt\}, minimalistic, geometric forms, constructivism art"
    &  23.50 & 96.31 & 72.81\\
    \midrule
    \multicolumn{1}{c}{Average} & - & 38.11& 79.91 & 41.80 \\
  \bottomrule
  \end{tabular}
  \vspace{-10pt}
\end{table}

\renewcommand\theadalign{cl}
\begin{table}[t]
  \caption{Samples of prompts for style image and plain image.
  }
  \label{tab:prompts}
  \centering
  \begin{tabular}{@{}ll@{}}
    \toprule
    \multicolumn{1}{c}{Prompt Samples for Style Image} \\
    \midrule
    
    "A man with a beard" \\
    "A lizard with red eyes" \\
    "A woman carrying a basket on her back" \\
    "Snowy night landscape with houses and trees in the snow" \\
    "A close up of a little hamster singing into a microphone" \\
    \toprule
    \multicolumn{1}{c}{Prompt Samples for Plain Image} \\
    \midrule
    "Affordable summer destinations hawaii" \\
    "Closeup of a spinach and feta cheese omelet" \\
    "Sportsman boxer fighting on black background with smoke boxing" \\
    "Healthy green smoothie and ingredients - detox and diet for health" \\
    "Masaru Kondo collects Ralph Lauren clothing and accessories." \\
  \bottomrule
  \end{tabular}
  \vspace{-10pt}
\end{table}

\renewcommand\theadalign{cl}
\begin{table}[t]
  \caption{Samples of text- and exemplar-based instruction templates. We utilize LLM to generate 15 templates for each task. "<style>" and "<image>" are identifiers for textual style name and exemplar image respectively.
  }
  \label{tab:temp_text}
  \centering
  \begin{tabular}{@{}l@{}}
    \toprule
    \multicolumn{1}{c}{Samples of Text-Based Instruction Templates}\\
    \midrule
"Please edit this image to embody the characteristics of <style> style." \\
"Transform this image to reflect the distinct aesthetic of <style>." \\
"Can you infuse this image with the signature techniques representative of <style>?" \\
"Adjust the visual elements of this image to emulate the <style> style." \\
"Reinterpret this image through the artistic lens of <style>." \\
    \toprule
    \multicolumn{1}{c}{Samples of Exemplar-Based Instruction Templates}\\
    \midrule
"Please match the aesthetic of this image to that of <image>." \\
"Adjust the current image to mimic the visual style of <image>." \\
"Edit this photo so that it reflects the artistic style found in <image>." \\
"Transform this picture to be stylistically similar to <image>." \\
"Recreate the ambiance and look of <image> in this one." \\
  \bottomrule
  \end{tabular}
  \vspace{-10pt}
\end{table}

\subsection{Textual Instruction Templates.}
We list 5 samples of machine-generated textual instruction templates for text- and exemplar-based style editing in~\cref{tab:temp_text} respectively.  We utilize LLM to generate 15 templates per task. In these instruction templates, "<style>" and "<image>" are planted as identifiers for textual style name and exemplar image. During training, we randomly choose from these templates to keep the syntax diversity.

\subsection{Data Evaluation.} \label{sec:supp_eval}
In~\cref{tab:styles}, we list the top styles where the usability improve the most after one editing-tuning round. The usability metric is also based on CLIP-score, which is the same with that of usability filtering. As we explained before, comparing to the results of vanilla de-style, the average usability rate is increased dramatically from 38.11\% to 79.91\% by 41.80\%. It shows that our Iterative Style-Destyle Editing is effective for quality improvement of paired images. 
Additionally, we show more visual results and comparison between $BatchA$, $BatchB$ and $BatchA^1$, $BatchA^2$, $BatchB^1$ in~\cref{fig:dataset2supp}, from which we can visually observe the differences.

\section{Experiment Implementation Details.} \label{sec:exp_detail}
We utilize 8 NVIDIA A800-SXM4-80GB GPUs for our experiments. During inference, we implement classifier-free guidance for both image and text conditions. Following the recommendation in \cite{guidancerescale}, we also apply a re-scaling factor of 0.5 to the outcomes. For instruction-based style editing, we fine-tune the pre-trained model\cite{ip2p} for 5000 steps under 0.0001 learning rate. For exemplar-based style editing, we only tune the alignment layers $W$ and the U-Net\cite{unet} decoder under the learning rate of 0.00001 for 35000 steps. Both are trained using Adam\cite{adamw} optimizer. The scale weighting is only applied during inference in compositional style editing tasks. We adjust scale factors in the range of $[0.5, 1.5]$.


\section{Additional Results.} \label{sec:supp_exp}

\begin{figure}[h]
  \centering
  \includegraphics[width=1.0\linewidth]{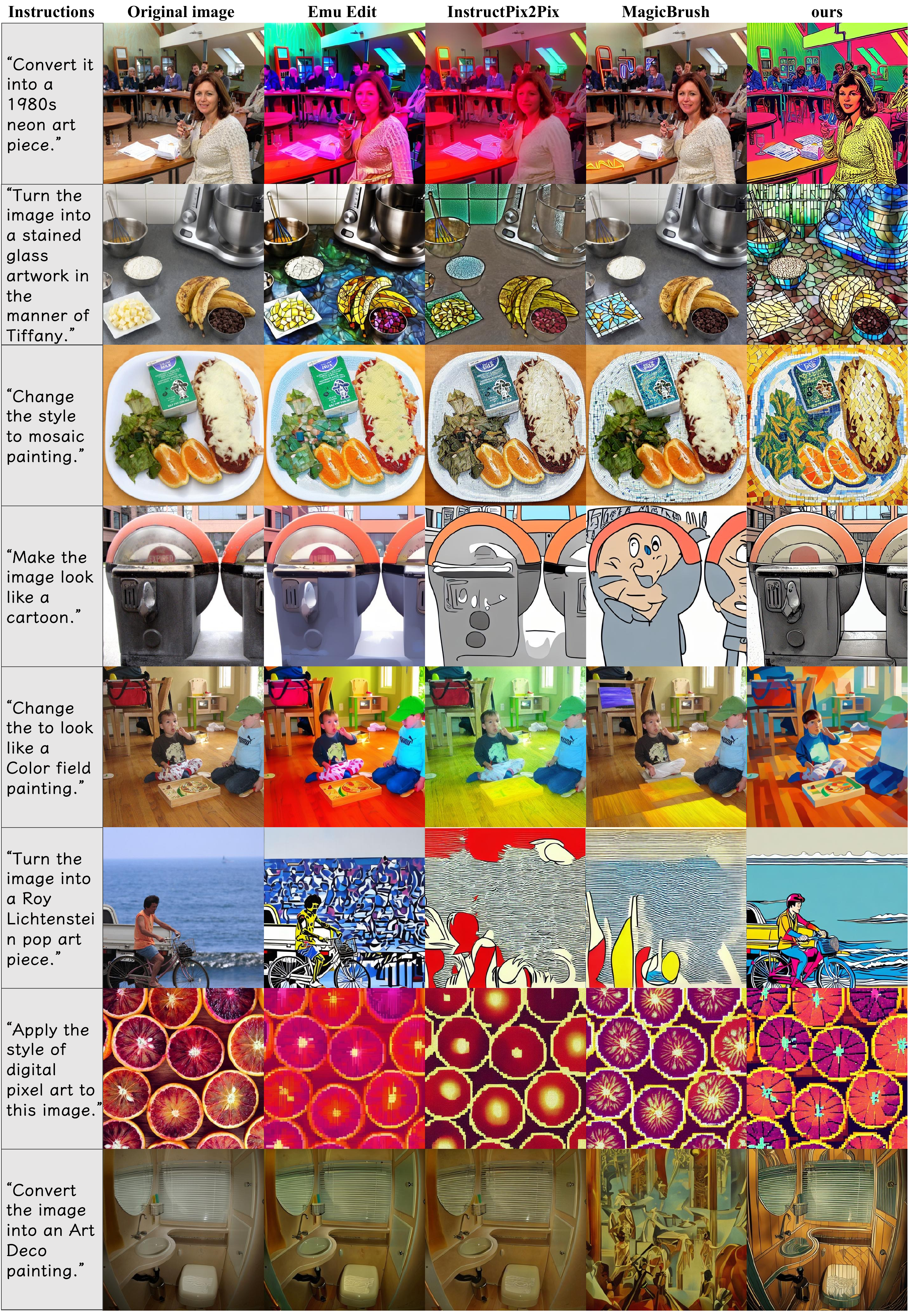}
  \caption{\textbf{Additional results comparing with baselines in Emu Edit benchmark.} 
  }
  \label{fig:comp1supp}
\end{figure}
\begin{figure}[tb]
  \centering
  \includegraphics[width=1.0\linewidth]{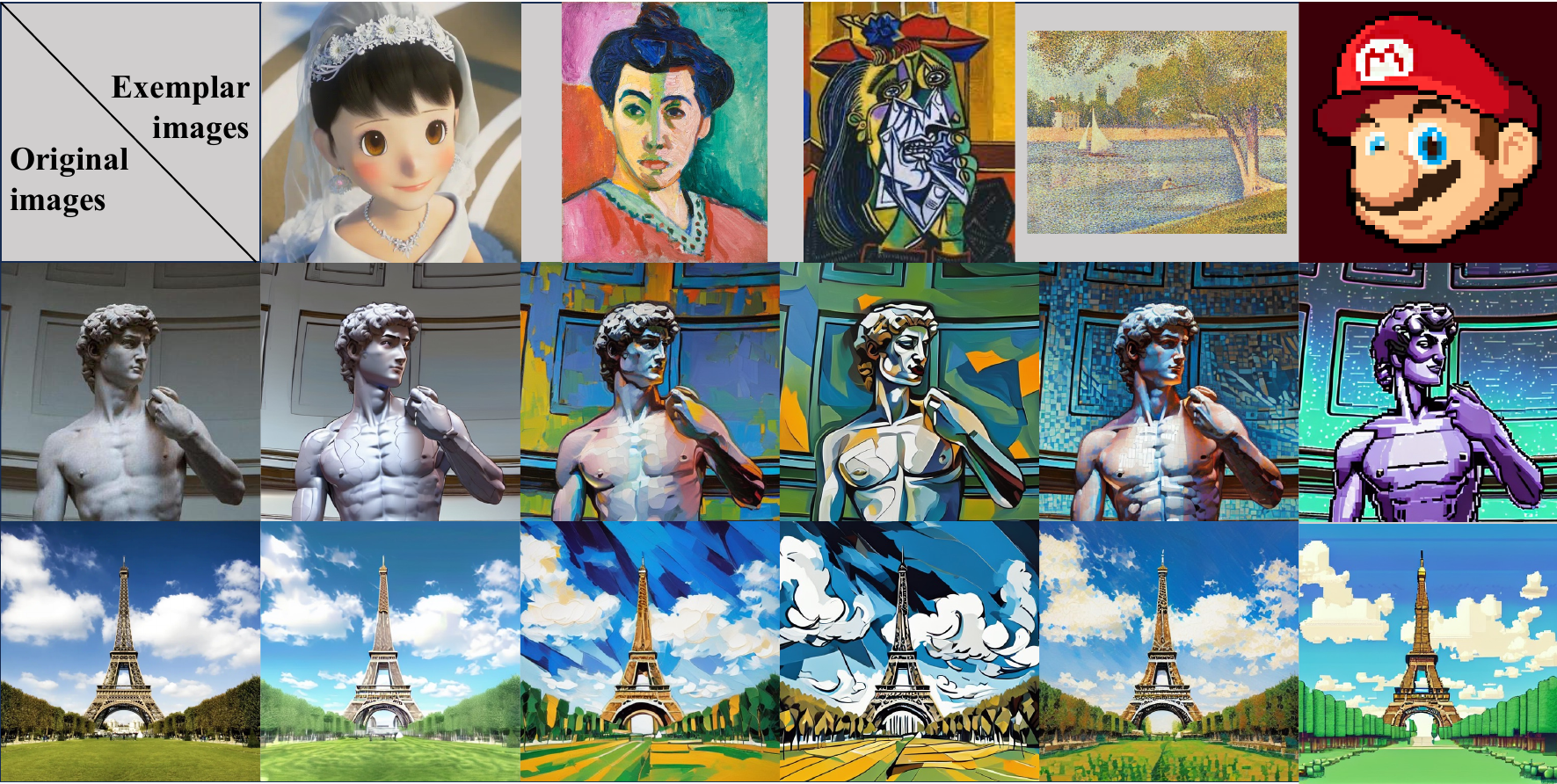}
  \caption{\textbf{Exemplar-based style editing with real world images.} We present the results of 2 original images in the styles of 5 different art works.
  }
  \label{fig:comp2supp}
\end{figure}
\begin{figure}[h]
  \centering
  \includegraphics[width=1.0\linewidth]{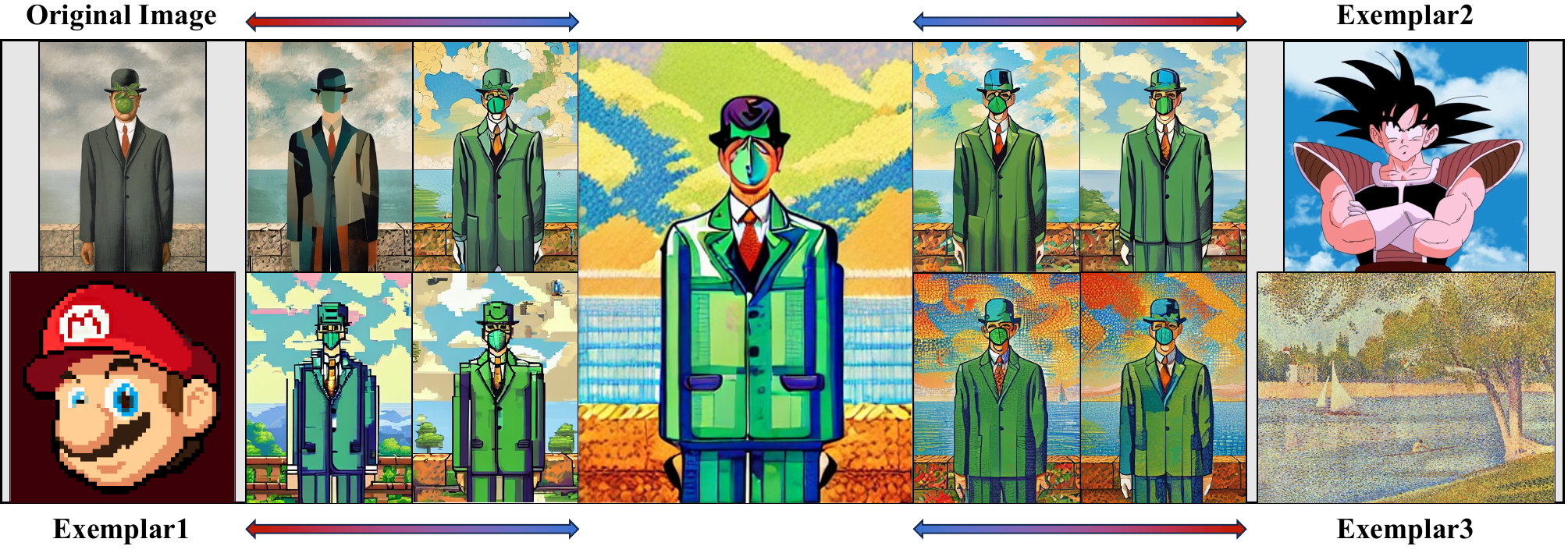}
  \caption{\textbf{Compositional style editing combining 3 different styles.} Both original image and the style exemplars are real world images.
  }
  \label{fig:comp3supp}
\end{figure}

As shown in~\cref{fig:comp1supp}, we present more qualitative results in Emu Edit benchmark, including comparisons with Emu Edit\cite{emuedit}, InstructPix2Pix\cite{ip2p} and Magic Brush\cite{magicbrush}. 
In~\cref{fig:comp2supp}, we show additional results of exemplar-based style editing with real world images. We use two original images: the David by Michelangelo and the Eiffel Tower, five style exemplars: an animate film stage photo, a Fauvism painting by Henri Matisse, a Cubism painting by Pablo Ruiz Picasso, a post-Impressionist painting by Georges Seurat and a pixel game character.
In~\cref{fig:comp3supp}, we demonstrate an editing result in a compositional style of 3 different styles. Both original image and the style exemplars are real world images. The art work "The Son of Man" by Rene Magritte is used as the original image and 3 exemplar images in different style are provided. We show the results under different scale weights.

\section{Discussions.} \label{sec:discus}
\subsection{Social Impacts.}

StyleBooth enables the facile editing of images through simple instructions, providing users with multimodal input options to manipulate images according to their preferences. 
This approach leads to some positive social impacts, as it allows users to achieve style transfer effortlessly without the need for professional editing tools, substantially enhancing productivity. 
Moreover, users can explore more liberating style options using text prompts or reference images, thereby offering a creatively enriched editing tool. 
However, given the inherent risks associated with generative image synthesis, such as malicious use and dissemination, it is imperative to incorporate additional safeguards during the development of such systematic tools.

\subsection{License of Assets.}
For baselines, Instruction-Pix2Pix\cite{ip2p} inherits this license as it is built upon Stable Diffusion. Magic Brush\cite{magicbrush} are released under Creative Commons Attribution 4.0 License. RIVAL\cite{rival}, StyleAligned\cite{stylealigned} and VCT\cite{vct} are under Apache-2.0 license.

For datasets, Emu Edit\cite{emuedit} and LAION Art\cite{laion5b} are under Creative Commons Attribution 4.0 License. According to Stable Diffusion-XL\cite{sdxl}, which is under Open RAIL++-M License, we have the right to distribute the generated images for research purpose.

\section{Human Evaluation.}\label{sec:human_eval}

\begin{figure}[h]
  \centering
  \includegraphics[width=1.0\linewidth]{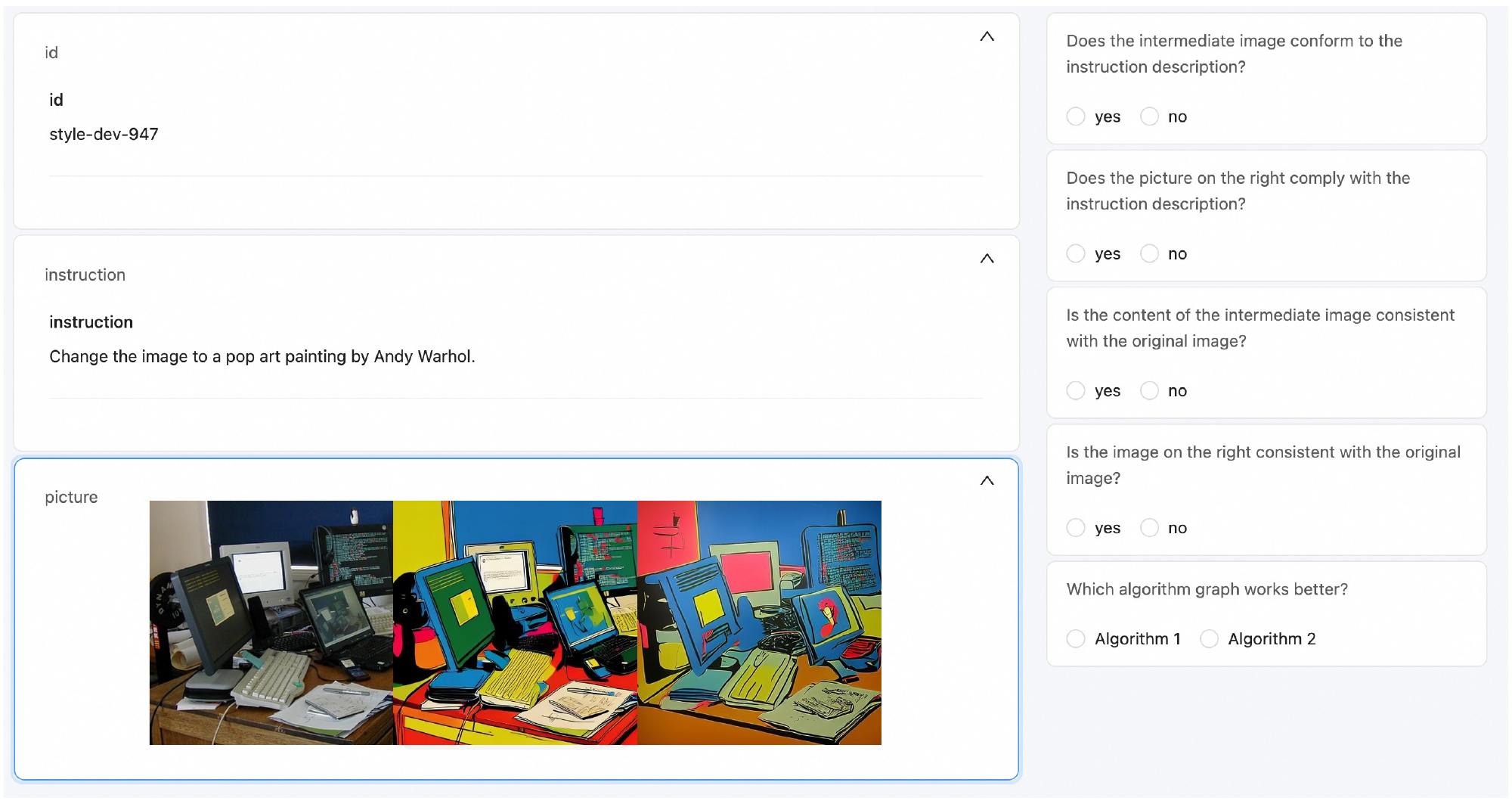}
  \caption{\textbf{Screenshots of our anonymous questionnaires for human evaluation.}}
  \label{fig:tag}
\end{figure}

We conducted human evaluation to assess the baseline comparisons. We distributed anonymous questionnaires with 5 questions for each editing case, along with instruction, original image, edit result from Method 1 and Method 2 as showed in \cref{fig:tag}. The first 2 questions are about whether styles of the 2 editing results are correct and match instructions. The third and fourth questions estimate the content consistency of the 2 methods. At last, the candidates were asked to choice the better one between these 2 methods. 

\clearpage 

\end{document}